\documentclass[letterpaper, 10 pt, conference]{ieeeconf}  

\IEEEoverridecommandlockouts                              

\usepackage{graphicx,verbatim}
\usepackage{multicol}
\usepackage{multirow}
\usepackage{booktabs}
\usepackage{amsmath}
\usepackage{hyperref}
\usepackage{amssymb}
\usepackage[table,xcdraw]{xcolor}

\title{\LARGE \bf
EndoMUST: Monocular Depth Estimation for Robotic Endoscopy via End-to-end Multi-step Self-supervised Training
}

\author{Liangjing Shao$^{1}$, Linxin Bai$^{1}$, Chenkang Du$^{1}$ and Xinrong Chen$^{1,\#}$
\thanks{This work was supported by National Natural Science Foundation of China (Grant No.82472116) and  Natural Science Foundation of Shanghai (Grant No.24ZR1404100).}
\thanks{$^{1}$Liangjing Shao, Linxin Bai, Chenkang Du and Xinrong Chen are with Academy for Engineering \& Technology, Fudan University, and also with Shanghai Key Laboratory of Medical Image Computing and Computer Assisted Intervention, Fudan University.}%
\thanks{\#: Corresponding Author (email: chenxinrong@fudan.edu.cn)}%
}

\begin{document}

\maketitle
\thispagestyle{empty}
\pagestyle{empty}

\begin{abstract}
Monocular depth estimation and ego-motion estimation are significant tasks for scene perception and navigation in stable, accurate and efficient robot-assisted endoscopy. To tackle lighting variations and sparse textures in endoscopic scenes, multiple techniques including optical flow, appearance flow and intrinsic image decomposition have been introduced into the existing methods. However, the effective training strategy for multiple modules are still critical to deal with both illumination issues and information interference for self-supervised depth estimation in endoscopy. Therefore, a novel framework with multistep efficient finetuning is proposed in this work. In each epoch of end-to-end training, the process is divided into three steps, including optical flow registration, multiscale image decomposition and multiple transformation alignments. At each step, only the related networks are trained without interference of irrelevant information. Based on parameter-efficient finetuning on the foundation model, the proposed method achieves state-of-the-art performance on self-supervised depth estimation on SCARED dataset and zero-shot depth estimation on Hamlyn dataset, with 4\%$\sim$10\% lower error. The evaluation code of this work has been published on https://github.com/BaymaxShao/EndoMUST.

\end{abstract}

\section{INTRODUCTION}
Endoscopy has been one of the most important minimally invasive surgical procedures. Robotic system could highly enhance the stability and accuracy of endoscopy. Due to the limited visual field and 2D vision of the endoscope, 3D scene perception is a critical task for more efficient, safe and accurate robot-assisted endoscopy. Depth estimation has been a common technique for 3D perception, which can also served as a basis of 3D reconstruction, registration and AR screening.\cite{imp, tmi}

The traditional methods for depth estimation are mainly based on feature matching from multi-view images.\cite{fm} Due to sparse feature in endoscopic images, the performance of traditional stereo matching methods is severely limited in endoscopy.\cite{fm_no} With high performance of feature representation, stereo matching based on deep learning has been proposed for depth estimation in endoscopy. Moreover, the inference speed of deep learning-based stereo matching methods cannot achieve the real-time requirements. Based on the most recent works \cite{sm}, the highest inference speed of the deep learning-based stereo matching is only 11 frames per second \cite{cgi} in endoscopic scenes. Thus, instead of traditional methods and binocular stereo matching, the deep learning-based monocular depth estimation methods with higher efficiency have gained common attention with the development of large-scale datasets. 

\begin{figure}
    \centering
    \includegraphics[width=8cm]{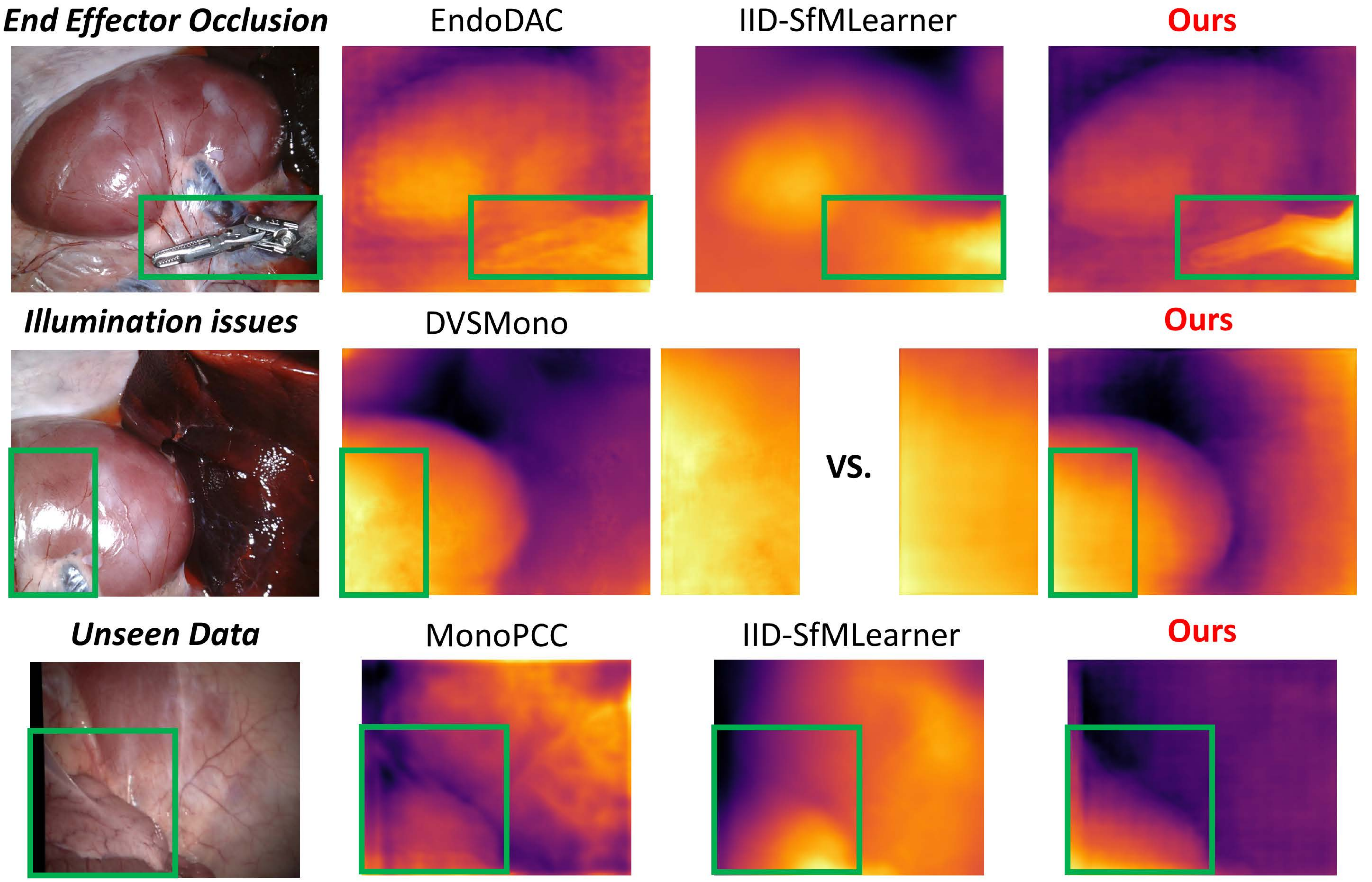}
    \caption{Challenges for depth estimation in robotic endoscopy. Highlights of our method are marked by green boxes.}
    \label{tease}
\end{figure}

Deep learning-based methods can be categorized into supervised depth estimation and self-supervised depth estimation. However, fully-supervised depth estimation methods, such as \cite{sde} and \cite{sde2}, will face the high cost for obtaining the ground truth of depth maps, especially for the narrow and enclosed space in endoscopy. Therefore, Mahmood et al.\cite{sde3} and Chen et al.\cite{sde4} utilize synthetic data to overcome the unavailability of large-scale and accurate RGB-D dataset for endoscopy. Nonetheless, the large domain gap between synthetic data and real data will make such methods fall short of desirable performance in realistic endoscopic scenes. Although a large quantity of self-supervised monocular depth estimation methods\cite{monovit,litemono,monodiff} have been proposed for natural scene, the depth estimation task is still challenging because of the lighting inconsistency and non-Lambertian reflection on soft tissues. In recent years, lots of researches have been proposed to tackle such problems in endoscopic scenes. Shao et al.\cite{afsfm} deal with the lighting inconsistency by generalized dynamic image constraint based on the appearance flow, while Yang et al.\cite{iidsfm} propose to filter the lighting reflectance by intrinsic image decomposition. However, the effective training strategy is still needed to efficiently and jointly tackle all illumination challenges in endoscopy. Meanwhile, for robotic endoscopy, the occlusion of the end effector is also a potential challenge for depth estimation.

In this work, a self-supervised framework with a novel multi-step training strategy is proposed to jointly cope with the challenges in robotic endoscopy shown as Fig. \ref{tease} with low cost. Specifically, each end-to-end training epoch consists of three training steps for different networks to prevent from the misleading of the irrelevant constraints. The contribution of this work can be summarized as the following:
\begin{enumerate}
    \item A novel three-step self-supervised training strategy is proposed to isolate different aspects of constraints and functions, with well-designed loss functions as well.
    \item In the second step, a novel multiscale self-supervised training method is proposed to capture more detail representations for optimization of intrinsic image decoposition. 
    \item In the third step, a full-module parameter-efficient finetuning based on DV-LoRA\cite{endodac} is applied into the training of depth map generation network, to control training cost and increase the adaptation.
    \item The proposed method achieves the state-of-the-art performance in both self-supervised depth estimation on SCARED dataset and zero-shot depth estimation on Hamlyn dataset, compared with existing methods.
    
\end{enumerate}

\section{RELATED WORK}

Self-supervised depth estimation has been developed for a long period. Turan et al.\cite{8593623} proposed the first method to estimate camera pose and depth information from image sequences using self-supervised learning for endoscopic capsule robots. Liu et al.\cite{liu2019dense} introduced the use of Structure from Motion (SfM) to extract supervision signals from monocular endoscopic videos. Li et al.\cite{li2020unsupervised} proposed to use peak signal-to-noise ratio to evaluate the similarity between synthesized and target images. In recent years, some works enhanced the network structure by utilizing attention mechanisms. Endo-SfMLearner\cite{endoslam} combined a spatial attention mechanism to highlight textured regions. Based on this, Liu et al.\cite{liu2023self} added position and channel attention modules to the network to capture multi-scale context, improving depth estimation accuracy. Furthermore, Yang et al.\cite{tmi} combined CNNs and Transformers in the depth estimation part to extract both local texture and global contour features. However, the issues of illumination and sparse texture are not considered in the above works.

In recent years, based on depth estimation for natural scenes, several works have focused on addressing illumination issues in endoscopic scenes. Shao et al.\cite{afsfm} introduced appearance flow, which accounts for changes in brightness patterns and incorporates the generalized dynamic image constraint. Differently, Li et al.\cite{iidsfm} proposed the Intrinsic Image Decomposition (IID) method, which decomposes an image into two fundamental components, albedo and shading for self-supervised learning. Based on the self-supervised training of \cite{afsfm}, some works begin to focus on the training strategy. With the development of foundation model for depth estimation\cite{da,da2}, EndoDAC\cite{endodac} proposed Dynamic Vector-Based Low-Rank Adaptation and convolutional neck blocks to efficiently finetune Depth Anything model\cite{da} for endoscopic depth estimation. Recently, DVSMono\cite{dvs} proposed a new dynamic view selection mechanism to address the pose alignment error from minimal camera motion. In this work, different from above works, a well-designed three-step training framework is proposed to jointly deal with feature gap, illumination inconsistency and reflection interference based on full-module parameter-efficient training.

\section{PROPOSED METHOD}

\subsection{Overview}

\begin{figure*}
    \centering
    \includegraphics[width=0.9\linewidth]{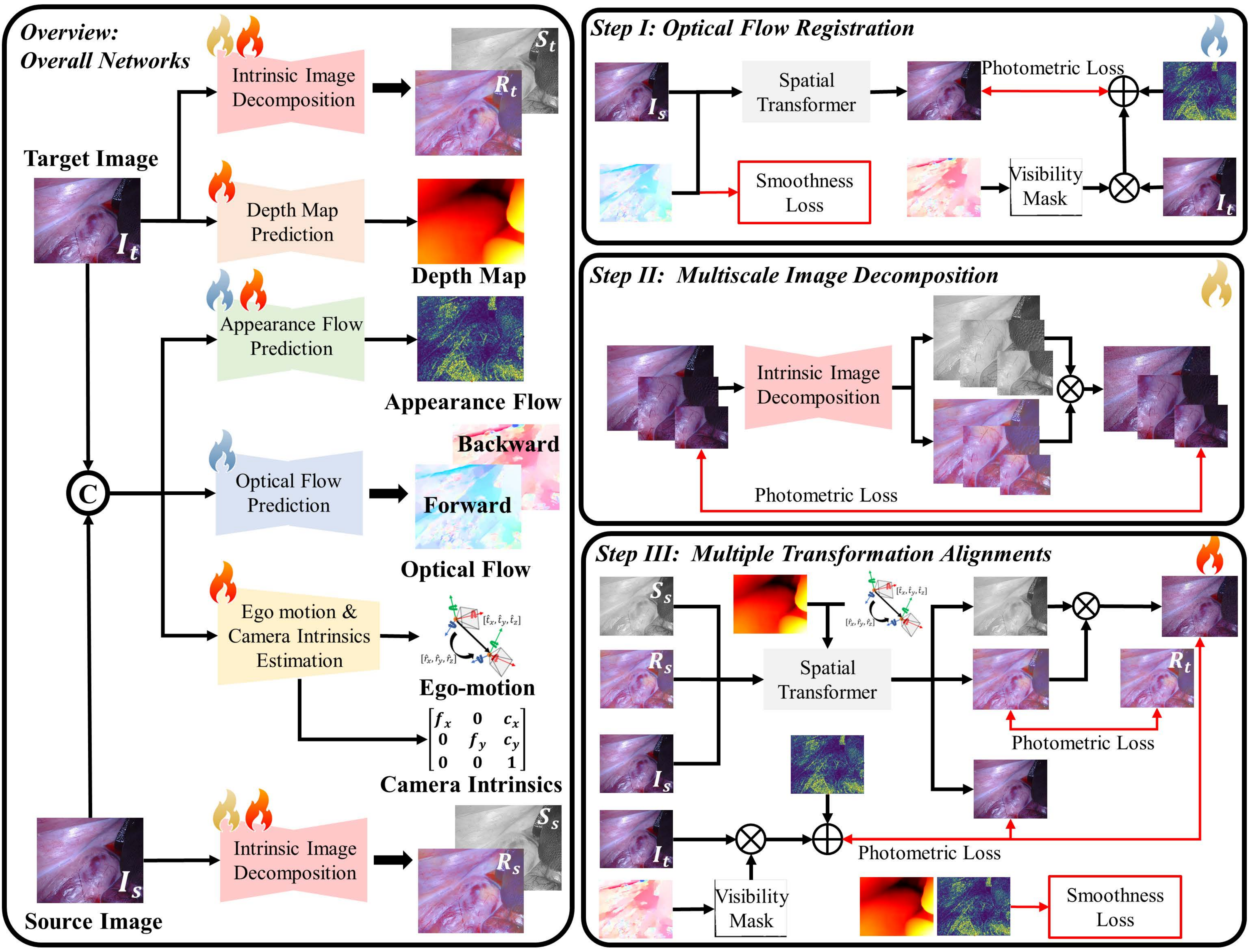}
    \caption{The overview of the proposed method. Each end-to-end training epoch is divided into three steps. In Step I, the optical flow prediction network and appearance flow prediction network are trained based on optical flow registration. In Step II, the intrinsic image decomposition network is trained based on self-supervision of reorganization. In Step III, all networks except optical flow prediction network are trained by the rigid transformation based on the predicted depth map, ego-motion and camera intrinsics.}
    \label{ov}
\end{figure*}

The overview of the proposed method is shown in Fig. \ref{ov}. Firstly, the appearance flow map $A$, forward optical flow map $O_f$, backward optical flow map $O_b$, ego-motion $T$ and camera intrinsics $K$ are separately generated by independent networks based on \cite{afsfm} and \cite{endodac} from the concatenation of target frame $I_t$ and source frame $I_s$. At the same time, using intrinsic image decomposition network from \cite{iidsfm}, the reflectance images and shading images of the target frame and the source frame are generated respectively. Meanwhile, the depth map $D_t$ of the target image is generated by the depth map generation network based on Depth Anything \cite{da}. Based on the generated images, in each training epoch, the training process is divided into three steps according to the different aspects of constraints. At each step, the related networks are trained while other networks are frozen. To generate the proper optical flow and appearance flow, the transformation alignment based on the optical flow is performed at the first step. Furthermore, for precise intrinsic image decomposition, the self-supervised learning on image decomposition network is performed based on multiscale source images at the second step. At the last step, the transformation alignments based on the predicted ego-motion and depth map are performed to train multiple related networks except optical flow generation network. The alignment losses are mainly based on photometric loss between the original image $I$ and the transformed image $\hat{I}$ defined as Eq. \ref{lp}
\begin{equation}
    \mathcal{L}_p(I,\hat{I})=(\alpha\frac{1-SSIM(I,\hat{I})}{2}+(1-\alpha)|I-\hat{I}|)
    \label{lp}
\end{equation}

\subsection{Step I: Optical Flow Registration}

To generate smooth optical flow map and appearance flow map, the smoothness loss of the forward optical flow map and the appearance flow map is calculated by Eq. \ref{l_so} and Eq. \ref{l_sa}, respectively. Based on the forward optical flow $O_f$, the source image can be transformed into the image $I^{opt}_{s\rightarrow t}$ via a differentiable inverse warping operation known as spatial transformer\cite{st}. The appearance difference based on Eq. \ref{opt_reg} is utilized to supervise the registration between the transformed image $I^{opt}_{s\rightarrow t}$ and the target image $I_t$. To this end, the loss function to train the optical flow estimation network and appearance flow estimation network at the first step can be defined as Eq. \ref{l1}.
\begin{equation}
    \mathcal{L}_{so} = |\nabla O_f| \cdot e^{-\nabla|O_f|}
    \label{l_so}
\end{equation}
\begin{equation}
    \mathcal{L}_{sa} = |\nabla A| \cdot e^{-\nabla|I_t-I^{opt}_{s\rightarrow t}|}
    \label{l_sa}
\end{equation}
\begin{equation}
    \mathcal{L}^{reg}_{opt} = \mathcal{L}_p(I^{opt}_{s\rightarrow t}, M\cdot I_t+A)
    \label{opt_reg}
\end{equation}
\begin{equation}
    \mathcal{L}_1=  \mathcal{L}^{reg}_{opt} + 0.001\mathcal{L}_{so} + 0.01\mathcal{L}_{sa}
    \label{l1}
\end{equation}

where $M$ denotes the visibility mask generated from the backward optical flow map based on \cite{vb}.

\subsection{Step II: Intrinsic Image Decomposition}
At this step, the source images and the target images $I_{\times1}$ will be scaled to images $I_{\times0.75}, I_{\times0.5}$ at smaller scales firstly. The multiscale source images and target images $\textbf{I}=\{I_{\times1}, I_{\times0.75}, I_{\times0.5}\}$ are decomposed into corresponding reflectance images $\textbf{R}$ and shading images $\textbf{S}$. Based on this, a reconstructed image $\hat{I}_i$ can be generated by multiplying each pair of reflectance image and shading image $(R_i\in\textbf{R}, S_i\in\textbf{S})$. For the alignment between decomposed images and corresponding input image $I_i\in\textbf{I}$, the appearance difference between the reconstructed image and the input image is calculated as the loss. Furthermore, the loss function of this step can be defined as Eq. \ref{l2}, which is utilized to train the intrinsic image decomposition network. 

\begin{equation}
    \mathcal{L}_2 = \frac{1}{3}\sum_i\mathcal{L}_p(I_i, R_i\times S_i)
    \label{l2}
\end{equation}

\subsection{Step III: Multiple Transformation Alignments}
At the beginning of the third step, the smoothness of the depth map and the appearance flow map is supervised by the function Eq. \ref{sm}.
\begin{equation}
    \mathcal{L}_{sm}=|\nabla A| \cdot e^{\nabla|I_t - I^{op}_{s\rightarrow t}|} + |\nabla D_t| \cdot e^{\nabla|I_t|}
    \label{sm}
\end{equation}

Given the depth map of the target image $D_t\in\mathbb{R}^{H\times W\times1}$, the ego-motion $P=[P_R\in\mathbb{R}^{3\times 3},P_t\in\mathbb{R}^{3\times1}]$ and camera intrinsics $K\in\mathbb{R}^{3\times3}$, the rigidly transformed image $I^{rg}_{s\rightarrow t}$ from the source image can be generated from Eq. \ref{rf}. In this equation, $I_s(p)$ and $I_{s\rightarrow t}(p)$ are the homogeneous matrix of corresponding pixel $p$ in the source image and the transformed image, while $\mathcal{C}(\cdot)$ denotes the coordination values in the homogeneous matrix. In addition, $P_R$ and $P_t$ are rotation matrix and translation vector of the ego-motion, respectively.
\begin{equation}
    [\mathcal{C}(K\cdot I_s(p)), D_s(p)]=P_R^{-1}([\mathcal{C}(K\cdot I_{s\rightarrow t}(p)), D_t(p)]-P_t)
    \label{rf}
\end{equation}
Similarly, the reflectance image $R_s$ and the shading image $S_s$ decomposed from the source image can also be transformed into the images $R_{s\rightarrow t}$ and $S_{s\rightarrow t}$. Additionally, the predicted target image $I^{iid}_{s\rightarrow t}=R_{s\rightarrow t}\times S_{s\rightarrow t}$ can also be composed by the transformed intrinsic images $R_{s\rightarrow t}$ and $S_{s\rightarrow t}$.

To supervise the depth map generation, the rigid transformation should make the transformed image and the target image aligned. Therefore, as Eq. \ref{tr} shows, the appearance differences between the target image and the transformed images, which are generated from the source image and its intrinsic images are applied for self-supervision. 

\begin{equation}
    \mathcal{L}_{tr}=\mathcal{L}_p(I^{iid}_{s\rightarrow t}, M\cdot I_t+A)+\mathcal{L}_p(I^{rg}_{s\rightarrow t},M\cdot I_t+A)
    \label{tr}
\end{equation}

In the reflectance image, the illumination influence has been filtered. Therefore, to alleviate the lighting interference, the illumination-free alignment is calculated based on the rigidly transformed reflectance image $R_{s\rightarrow t}$ and the reflectance image $R_t$ of the target image by Eq. \ref{iia}.

\begin{equation}
    \mathcal{L}_{iia}=\mathcal{L}_p(R_{s\rightarrow t}, M\cdot R_t)
    \label{iia}
\end{equation}

Finally, the loss function for the last step can be defined as Eq. \ref{l3}.
\begin{equation}
    \mathcal{L}_3=0.001\mathcal{L}_{sm}+0.01\mathcal{L}_{tr}+0.02\mathcal{L}_{iia}
    \label{l3}
\end{equation}

\subsection{Full-module Parameter-efficient Finetuning}
\begin{figure}
    \centering
    \includegraphics[width=6cm]{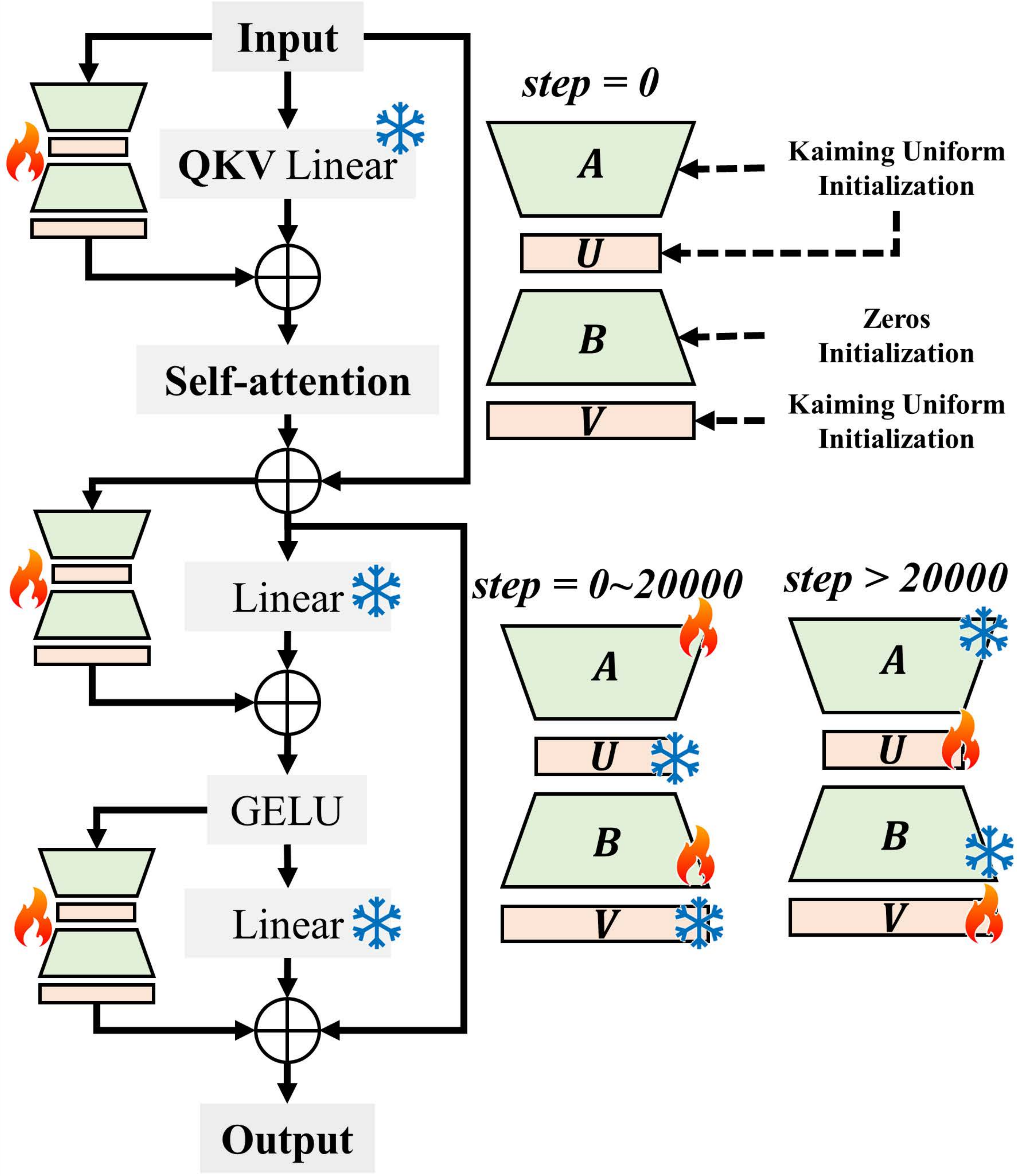}
    \caption{Parameter-efficient finetuning in each Transformer block based on DV-LoRA\cite{endodac}.}
    \label{ft}
\end{figure}
At the third step, full-module parametwr-efficient finetuning is performed on the Transformer blocks in depth map generation network as Fig. \ref{ft} displays. For each Transformer block in the encoder of depth map generation network, two linear layers in the feed forward network and the linear layer, which generates query, key and value vector, are finetuned by DV-LoRA\cite{endodac}. In detail, to finetune the module denoted as $\mathcal{F}$, given the input $X$, the output $Y=\mathcal{F}(X)$ will be adjusted by Eq. \ref{dvlora}.
\begin{equation}
    Y'=Y+\Lambda_VB\Lambda_UAX
    \label{dvlora}
\end{equation}
where $\Lambda_V$ and $\Lambda_U$ diagonal matrix of trainable weights $U$ and $V$, while $Y$ is frozen and $A,B$ are also trainable weights.

\section{EXPERIMENTS AND RESULTS}

\begin{table*}[]
\centering
\caption{Results of Depth Estimation on SCARED Dataset}
\begin{tabular}{c|c|ccccc|cc}
\toprule
\textbf{Methods}                     & \textbf{GC} & \textbf{$Rel_{Abs}\downarrow$} & \textbf{$Rel_{Sq}\downarrow$} & \textbf{$RMSE\downarrow$} & \textbf{$RMSE_{Log}\downarrow$} & \textbf{$\delta\uparrow$} & \textbf{TP/M} & \textbf{Speed/ms} \\ \midrule
AF-SfMLearner \cite{afsfm}            & Yes         & 0.059                          & 0.435                         & 4.925                     & 0.082                           & 0.974                     & 14.8          & 2.0               \\
Yang et al. \cite{tmi}                & Yes         & 0.062                          & 0.558                         & 5.585                     & 0.090                           & 0.962                     & 2.0           & 8.0               \\
Depth Anything$\dagger$ \cite{da}     & Yes         & 0.055                          & 0.410                         & 4.769                     & 0.078                           & 0.973                     & 13.0          & 5.0               \\
Depth Anything v2$\dagger$ \cite{da2} & Yes         & 0.076                          & 0.683                         & 6.379                     & 0.104                           & 0.949                     & 13.0          & 5.0               \\
IID-SfMLearner \cite{iidsfm}          & Yes         & 0.057                          & 0.430                         & 4.822                     & 0.079                           & 0.972                     & 14.8          & 2.0               \\
DVSMono \cite{dvs}                    & Yes         & 0.055                          & 0.410                         & 4.797                     & 0.078                           & 0.975                     & 27.0          & 12.7              \\
MonoPCC \cite{pcc}                    & Yes         & \underline{0.051}                          & \underline{0.349}                         & \underline{4.488}                     & \underline{0.072}                           & \underline{0.983}                     & 27.0          & 12.6              \\
EndoDAC \cite{endodac}                & No          & \underline{0.051}                          & 0.365                         & 4.545                     & \underline{0.072}                           & 0.982                     & 1.6           & 5.7               \\
EndoMUST(Ours)                            & No          & \textbf{0.046}                 & \textbf{0.313}                & \textbf{4.276}            & \textbf{0.067}                  & \textbf{0.984}            & 1.8           & 6.2               \\ \bottomrule
\multicolumn{9}{l}{\textbf{TP} denotes the number of Trainable Parameters in the Depth Net, \textbf{GC} denotes Given Camera intrinsics.}\\
\multicolumn{9}{l}{$\dagger$: Finetuned on SCARED dataset based on EndoDAC\cite{endodac}}
\end{tabular}
\label{res1}
\end{table*}

\begin{table*}[]
\centering
\caption{Results of Zero-shot Depth Estimation on Hamlyn Dataset}
\begin{tabular}{ccccccccc}
\toprule
\multicolumn{1}{c|}{\textbf{Methods}}                     & \multicolumn{1}{c|}{\textbf{GC}} & \textbf{$Rel_{Abs}\downarrow$} & \textbf{$Rel_{Sq}\downarrow$} & \textbf{$RMSE\downarrow$} & \textbf{$RMSE_{Log}\downarrow$} & \multicolumn{1}{c|}{\textbf{$\delta\uparrow$}} & \textbf{TP/M} & \textbf{Speed/ms} \\ \midrule
\multicolumn{1}{c|}{Endo Depth \& Motion \cite{edm}}        & \multicolumn{1}{c|}{Yes}         & 0.185                          & 5.424                         & 16.100                    & 0.225                           & \multicolumn{1}{c|}{0.732}                     & -             & $\sim$15               \\
\multicolumn{1}{c|}{AF-SfMLearner \cite{afsfm}}            & \multicolumn{1}{c|}{Yes}         & 0.168                          & 4.440                         & 13.870                    & 0.204                           & \multicolumn{1}{c|}{0.770}                     & 14.8          & 1.0               \\
\multicolumn{1}{c|}{Depth Anything$\dagger$ \cite{da}}     & \multicolumn{1}{c|}{Yes}         & 0.154                          & 3.616                         & 12.733                    & 0.189                           & \multicolumn{1}{c|}{0.784}                     & 13.0          & 3.3               \\
\multicolumn{1}{c|}{Depth Anything v2$\dagger$ \cite{da2}} & \multicolumn{1}{c|}{Yes}         & 0.182                          & 4.994                         & 15.067                    & 0.219                           & \multicolumn{1}{c|}{0.740}                     & 13.0          & 3.3               \\
\multicolumn{1}{c|}{IID-SfMLearner \cite{iidsfm}}          & \multicolumn{1}{c|}{Yes}         & 0.171                          & 4.526                         & 14.066                    & 0.206                           & \multicolumn{1}{c|}{0.767}                     & 14.8          & 1.0               \\
\multicolumn{1}{c|}{DVSMono \cite{dvs}}                    & \multicolumn{1}{c|}{Yes}         & \underline{0.152}                          & \underline{3.566}                         & \underline{12.64}3                    & \underline{0.187}                           & \multicolumn{1}{c|}{0.791}                     & 27.0          & 10.5              \\
\multicolumn{1}{c|}{MonoPCC \cite{pcc}}                    & \multicolumn{1}{c|}{Yes}         & 0.158                          & 3.889                         & 13.205                    & 0.194                           & \multicolumn{1}{c|}{0.782}                     & 27.0          & 10.5              \\
\multicolumn{1}{c|}{EndoDAC \cite{endodac}}                & \multicolumn{1}{c|}{No}          & 0.153                          & 3.655                         & 12.668                    & 0.188                           & \multicolumn{1}{c|}{\underline{0.795}}                     & 1.6           & 4.0               \\
\multicolumn{1}{c|}{EndoMUST(Ours)}                       & \multicolumn{1}{c|}{No}          &  \textbf{0.145}                 & \textbf{3.382}                & \textbf{12.100}           & \textbf{0.180}                  & \multicolumn{1}{c|}{\textbf{0.807}}            & 1.8           & 4.5               \\ \bottomrule
\multicolumn{9}{l}{\textbf{TP} denotes the number of Trainable Parameters in the Depth Net, \textbf{GC} denotes Given Camera intrinsics.}\\
\multicolumn{9}{l}{$\dagger$: Finetuned on SCARED dataset based on EndoDAC\cite{endodac}}
\end{tabular}
\label{res2}
\end{table*}

\begin{figure*}
    \centering
    \includegraphics[width=\linewidth]{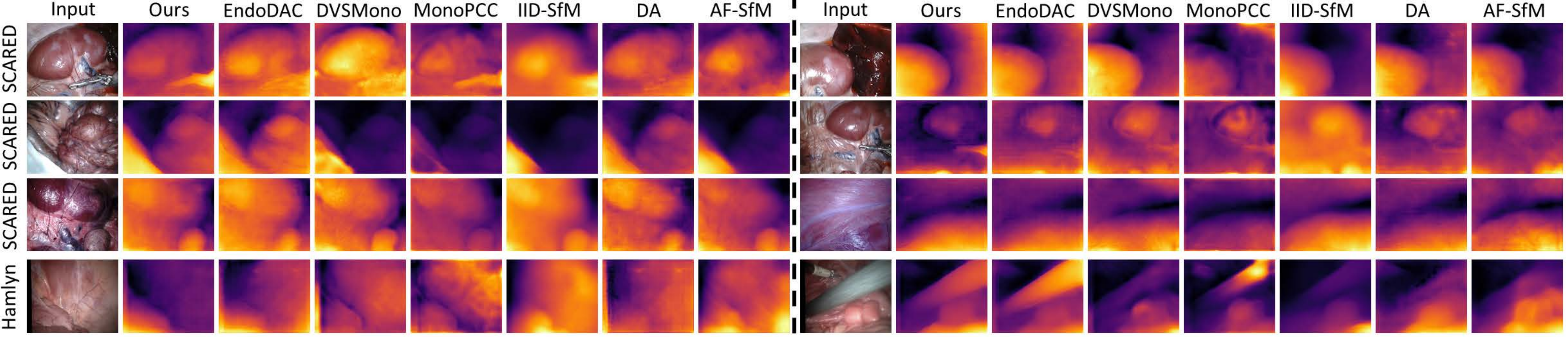}
    \caption{Qualitative Results of Depth Estimation. Results of self-supervised depth estimation on SCARED dataset are in first three rows. Results of zero-shot depth estimation on Hamlyn dataset are in the last row.}
    \label{vis-depth}
\end{figure*}

\subsection{Datasets}
\subsubsection{SCARED} SCARED dataset\cite{scared} consists of 35 endoscopic videos collected from fresh pig cadaver abdominal samples using the da Vinci surgical robotic system. The resolution of each image in this dataset is 1280×1024, which is adjusted to 320×256 in the experiment. Following AF-SfMLearner\cite{afsfm}, this dataset is divided into 15351, 1705, and 551 frames, which are used as the training, validation, and test sets in the depth estimation experiment, respectively.
\subsubsection{Hamlyn} Hamlyn\cite{hamlyn} dataset contains vast endoscopic videos acquired by the da Vinci surgical robot system during various surgical procedures. The images in this dataset have multiple resolutions, and in the experiments, all image resolutions are uniformly adjusted to 320 × 256. In the experiments, all 92,672 endoscopic frames from the 21 rectified videos\cite{edm} in this dataset are used to evaluate the zero-shot depth estimation.
\subsection{Experiments Implementation}
The proposed framework is implemented on one NVIDIA RTX 4090 GPU with Pytorch. The rank of DV-LoRA is set to 4, while the warm-up step is set to 20000. For each training step, an Adam optimizer is set with initial learning rate of 1e-4, which is scaled by a factor of 0.1 after 10 epochs. The framework is trained for 20 epochs in total with a batch size of 8.
\subsection{Evaluation Metrics}
Following previous works, four error metrics including Absolute Relative Error ($Rel_{Abs}$), Squared Absolute Relative Error ($Rel_{Sq}$), Root Mean Square Error ($RMSE$) and its logarithmic term ($RMSE_{Log}$) are used as evaluation metrics for depth estimation, as well as the $\delta$ to evaluate the accuracy, which are defined as the following equations, where $\mathbf{D}$ is the set of the predicted depth. $\hat{d}$ and $d$ denote the predicted depth and the ground truth for each pixel.
\begin{equation}
    Rel_{Sq}=\frac{1}{|\mathbf{D}|}\sum_{\hat{d}\in\mathbf{D}}\frac{|d-\hat{d}|^{2}}{d}
\end{equation}
\begin{equation}
    RMSE=\sqrt{\frac{1}{|\mathbf{D}|}\sum_{\hat{d}\in\mathbf{D}}|\log d-\log \hat{d}|^2}
\end{equation}
\begin{equation}
    RMSE_{Log} = \sqrt{\frac{1}{|\mathbf{D}|}\sum_{\hat{d}\in\mathbf{D}}|d-\hat{d}|^2}
\end{equation}
\begin{equation}
    \delta=\frac{1}{|\mathbf{D}|}\left|\left\{\hat{d}\in\mathbf{D}|max(\frac{d}{\hat{d}},\frac{\hat{d}}{d}<1.25)\right\}\right|
\end{equation}

The Absolute Trajectory Error (ATE) in Eq. \ref{ate} is utilized to evaluate ego-motion estimation, while Percentage Relative Error (PRE)  in Eq. \ref{pre} is used to evaluate camera intrinsics estimation. 
\begin{equation}
    ATE=\left\lVert P-\hat{P} \right\rVert
    \label{ate}
\end{equation}
\begin{equation}
    PRE=\frac{|k-\hat{k}|}{k}\times100\%
    \label{pre}
\end{equation}
where $\hat{k}$ and $k$ denote the predicted result and the ground truth of each element in camera intrinsics, $\hat{P}$ and $P$ denote trajectory point positions from the predicted results and the ground truths of ego-motion.

\subsection{Depth Estimation Results}
The proposed method is compared with state-of-the-art methods for depth estimation in endoscopy, including Endo Depth \& Motion \cite{edm}, AF-SfMLearner \cite{afsfm}, Yang et al. \cite{tmi}, IID-SfMLearner \cite{iidsfm}, DVSMono \cite{dvs}, MonoPCC \cite{pcc}, EndoDAC \cite{endodac}, as well as finetuned foundation model Depth Anything \cite{da,da2}. Due to differences in the hardware and dataset splits, some works with public codes or public weights are trained and tested based on the same setting in this work.

\subsubsection{Self-supervised Depth Estimation on SCARED Dataset} As Table \ref{res1} shows, the proposed method outperforms existing methods on SCARED dataset, with 9.23\% lower error on average and higher accuracy according to $\delta$. At the same time, the training cost of this work is just a little higher than EndoDAC\cite{endodac} with more 0.2M trainable parameters, while lower than other works. The inference speed of this work is lower than most of previous works, but the speed of over 100 fps can still achieve the real-time requirement. Qualitative results in Fig further display the outstanding performance of the proposed  method. Qualitative results are also displayed in Fig. \ref{vis-depth}.

\subsubsection{Zero-shot Depth Estimation on Hamlyn Dataset} The models with weights from training on SCARED dataset are utilized for evaluation on Hamlyn dataset. As \cite{tmi} and \cite{edm} are without public code, the proposed method is compared with \cite{edm} instead of \cite{tmi}, which is not compared on SCARED dataset. The experimental results in Table \ref{res2} demonstrate the best generalizability of the proposed method, with 4.45\% lower error on average and 1.51\% higher accuracy for zero-shot depth estimation on Hamlyn dataset. Moreover, the inference speed of the proposed model still achieves over 200 fps, although lower than some previous works. Qualitative results are also displayed in Fig. \ref{vis-depth}.

\begin{table}[]
\centering
\caption{Results of Ego-motion Estimation}
\begin{tabular}{c|ccc}
\hline
\multirow{2}{*}{\textbf{Methods}} & \multicolumn{3}{c}{\textbf{ATE}}                                \\ \cline{2-4} 
                                  & \textbf{Sequence 1} & \textbf{Sequence 2} & \textbf{Sequence 3} \\ \hline
AF-SfMLearner\cite{afsfm}         & 0.0941              & \textbf{0.0742}     & 0.0596              \\
IID-SfMLearner\cite{iidsfm}       & 0.0951              & 0.0764              & 0.0614              \\
MonoPCC\cite{pcc}                 & 0.1040              & 0.0781              & 0.0666              \\
EndoDAC\cite{endodac}             & \underline{0.0936}        & 0.0776              & \underline{0.0588}        \\
EndoMUST(Ours)                    & \textbf{0.0933}     & \underline{0.0750}        & \textbf{0.0586}     \\ \hline
\end{tabular}
\label{traj}
\end{table}

\begin{table}[]
\centering
\caption{Results of Camera Intrinsics Estimation}
\begin{tabular}{cc|cccc}
\hline
\multicolumn{2}{c|}{\textbf{Ground Truth}} & \textbf{fx}=0.82    & \textbf{fy}=1.02    & \textbf{cx}=0.5    & \textbf{cy}=0.5    \\ \hline
\multirow{2}{*}{EndoDAC\cite{endodac}} & Predicted & 0.8624         & 1.0786         & 0.4931         & 0.5091         \\
                                       & PRE       & 5.17\%          & 5.75\%          & 1.38\%          & 1.82\%          \\ \hline
\multirow{2}{*}{EndoMUST}              & Predicted & 0.8168         & 1.0230         & 0.4971         & 0.4964         \\
                                       & PRE       & \textbf{0.39\%} & \textbf{0.29\%} & \textbf{0.58\%} & \textbf{0.72\%} \\ \hline
\end{tabular}
\label{ci}
\end{table}

\begin{figure}
    \centering
    \includegraphics[width=8.5cm]{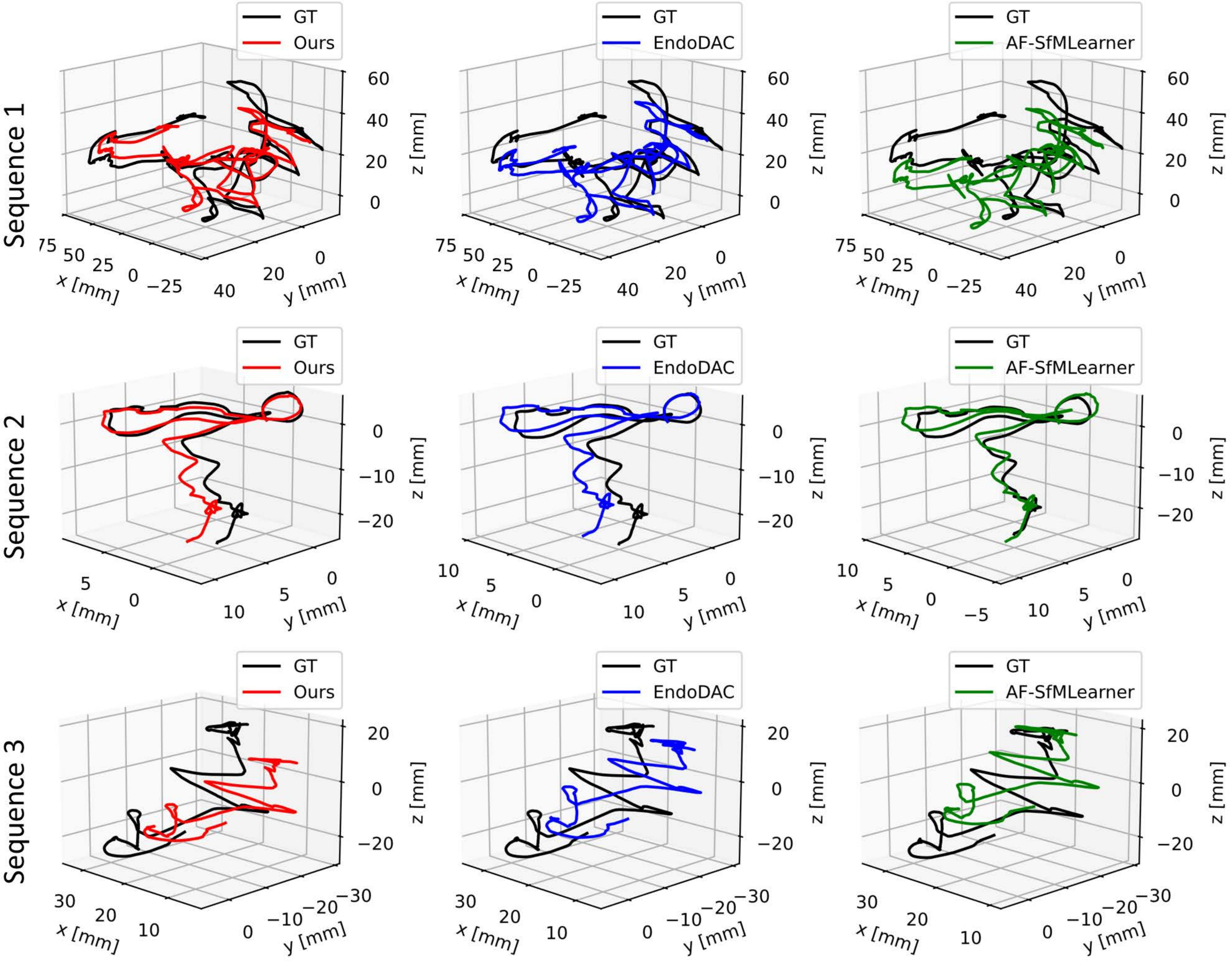}
    \caption{Qualitative results of ego-motion estimation.}
    \label{vis_traj}
\end{figure}

\subsection{Ego-motion and Camera Intrinsics Estimation Results}
From the test split of SCARED dataset, 3 sequences are selected for evaluation of ego-motion and camera intrinsics estimation. Based on a series of ego-motion between two continuous frames predicted by the network trained along with the framework at the third-step finetuning, the corresponding trajectory of each sequence can be calculated. Compared the predicted trajectories from our method and existing methods, results in Table \ref{traj} and Fig. \ref{vis_traj} demonstrate that the proposed model can provide the most accurate ego-motion estimation. Furthermore, on 7 test sequences of SCARED dataset, the camera intrinsics predicted by the proposed model and EndoDAC are also compared in Table \ref{ci}. The relative error of focal length estimation based on our method is 5\% lower than \cite{endodac}, while the relative error of optical center position estimation is 1.5$\times$ lower.

\begin{table}[]
\centering
\caption{Ablations on Training Steps and Strategies}
\begin{tabular}{c|ccccc}
\hline
\textbf{Steps} & \textbf{$Rel_{Abs}$} & \textbf{$Rel_{Sq}$} & \textbf{$RMSE$} & \textbf{$RMSE_{Log}$} & \textbf{$\delta$} \\ \hline
{\cellcolor[HTML]{EFEFEF}I$\rightarrow$II$\rightarrow$III}      & {\cellcolor[HTML]{EFEFEF}\textbf{0.046}}       & {\cellcolor[HTML]{EFEFEF}\textbf{0.313}}      & {\cellcolor[HTML]{EFEFEF}\textbf{4.276}}  & {\cellcolor[HTML]{EFEFEF}\textbf{0.067}}        & {\cellcolor[HTML]{EFEFEF}\textbf{0.984}}    \\
I$\rightarrow$II$\rightarrow$III$^\star$      & 0.050       & 0.352      & 4.478  & 0.071       & 0.983    \\
I$\rightarrow$II$^\dagger$$\rightarrow$III      & 0.049       & 0.344      & 4.441  & 0.070       & 0.983    \\
III$\rightarrow$I$\rightarrow$II      & 0.048               & 0.328               & 4.335         & 0.068                 & \textbf{0.984}            \\
I$\rightarrow$\{II,III\}    & 0.050                & 0.334               & 4.342          & 0.071                 & 0.983             \\
I$\rightarrow$\{II,III\}$^\ddagger$    & 0.050                & 0.335               & 4.358          & 0.070                 & 0.983             \\
\{I,II\}$\rightarrow$III    & 0.049                & 0.325              & 4.308         & 0.068                 & \textbf{0.984}             \\ 
I$\rightarrow$III    & 0.049                & 0.348              & 4.500         & 0.070                 & 0.983             \\ \hline
II$\rightarrow$\{I,III\}    & \multicolumn{5}{c}{\multirow{2}{*}{{Out of GPU Memory}}}               \\
\{I,II,III\}    & \multicolumn{5}{c}{}            \\  \hline
\multicolumn{6}{l}{'\{\}' denotes the steps are fused into one step.}\\
\multicolumn{6}{l}{$\star$: without finetuning QKV Linear in Fig.\ref{ft}.}\\
\multicolumn{6}{l}{$\dagger$: without multiscale image decomposition.}\\
\multicolumn{6}{l}{$\ddagger$: without intrinsic image decomposition.}
\end{tabular}
\label{ab1}
\end{table}

\begin{table}[]
\centering
\caption{Ablations on Warm-up Steps of DV-LoRA}
\begin{tabular}{c|ccccc}
\hline
\textbf{Warm-up} & \textbf{$Rel_{Abs}$} & \textbf{$Rel_{Sq}$} & \textbf{$RMSE$} & \textbf{$RMSE_{Log}$} & \textbf{$\delta$} \\ \hline
25000   & 0.048                & 0.320               & 4.303          & \textbf{0.067}                 & \textbf{0.985}   \\
{\cellcolor[HTML]{EFEFEF}20000}      & {\cellcolor[HTML]{EFEFEF}\textbf{0.046}}       & {\cellcolor[HTML]{EFEFEF}\textbf{0.313}}      & {\cellcolor[HTML]{EFEFEF}\textbf{4.276}}  & {\cellcolor[HTML]{EFEFEF}\textbf{0.067}}        & {\cellcolor[HTML]{EFEFEF}0.984}    \\
15000    & \textbf{0.046}                & 0.316              & 4.290          & \textbf{0.067}                 & 0.984   \\
5000    & 0.051                & 0.370               & 4.609          & 0.072                 & 0.983    \\ \hline
\end{tabular}
\label{warms}
\end{table}

\subsection{Ablation Studies}

\begin{figure*}
    \centering
    \includegraphics[width=0.9\linewidth]{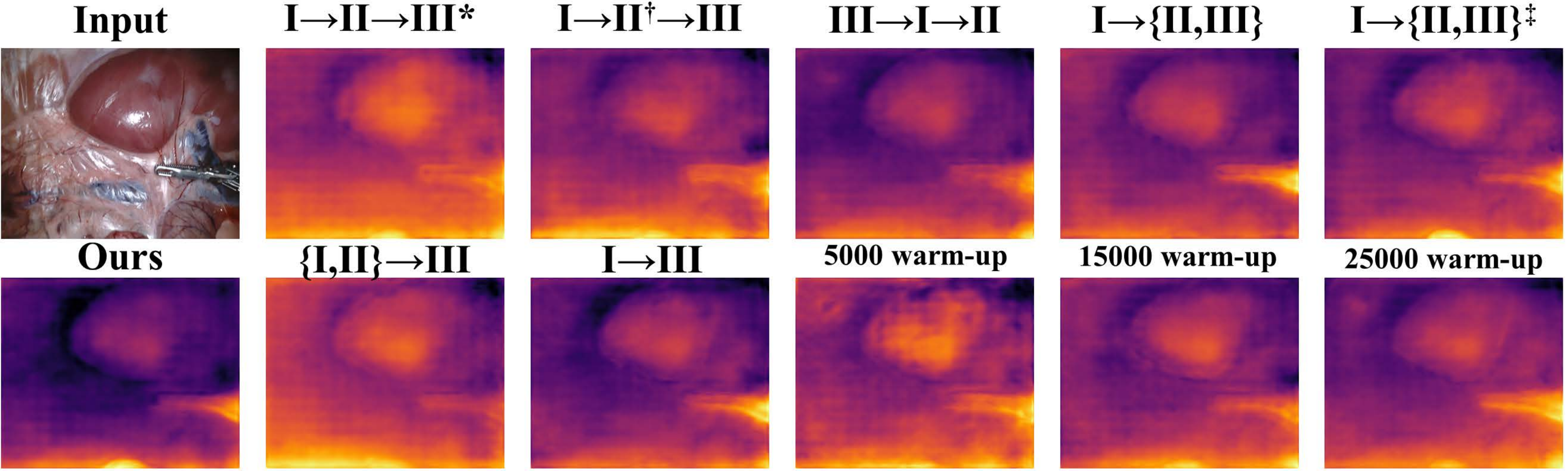}
    \caption{Qualitative Comparison of Ablation Studies on SCARED dataset.}
    \label{abs}
\end{figure*}

A series of ablation studies are conducted to demonstrate the effects of the proposed training strategy. Since steps I and II are independent of each other, their order won't affect the performance. The depth estimation results on SCARED dataset based on different orders and combinations of three steps are shown in Table \ref{ab1}, which demonstrates the outstanding performance based on the proposed setting. Note that to fuse the step I and step II, the loss function Eq. \ref{l12} is used to train the fused step based on Eq. \ref{l1} and Eq. \ref{l2}. Similarly, the loss function Eq. \ref{l23} is utilized to train the fusion of step II and step III.
\begin{equation}
    \mathcal{L}_{12}= \mathcal{L}_1+0.02\mathcal{L}_2
    \label{l12}
\end{equation}
\begin{equation}
    \mathcal{L}_{23}= 0.02\mathcal{L}_2+\mathcal{L}_3
    \label{l23}
\end{equation}
As the depth map generation network and optical flow generation network are only trained in stage I and stage III respectively, only stage II is removed to demonstrate the effect in ablation studies. Moreover, the effect from the full-module parameter-efficient finetuning including finetuning QKV layer is also proved based on the experiments. Meanwhile, the effects of intrinsic image decomposition and multiscale image decomposition supervision are demonstrated by several ablation studies as well. Furthermore, varied settings of warm-up steps are also evaluated in Table \ref{warms}. Besides, in Fig. \ref{abs}, qualitative results from different ablation studies are displayed.

\section{CONCLUSION}
In this work, a self-supervised depth estimation framework is proposed based on novel multi-step and parameter-efficient finetuning with intrinsic image decomposition. The proposed method not only achieves the state-of-the-art performance on self-supervised depth estimation, but also exhibits the best generalization performance for zero-shot depth estimation on over 90,000 frames. This work could contribute to more accurate, more efficient and more automatic robot-assisted minimally invasive surgery. Furthermore, feature diversity in various endoscopic scenes is still a challenge for generalization of the model.

\bibliographystyle{IEEEtran}
\bibliography{refs}

\begin{thebibliography}{10}
\providecommand{\url}[1]{#1}
\csname url@samestyle\endcsname
\providecommand{\newblock}{\relax}
\providecommand{\bibinfo}[2]{#2}
\providecommand{\BIBentrySTDinterwordspacing}{\spaceskip=0pt\relax}
\providecommand{\BIBentryALTinterwordstretchfactor}{4}
\providecommand{\BIBentryALTinterwordspacing}{\spaceskip=\fontdimen2\font plus
\BIBentryALTinterwordstretchfactor\fontdimen3\font minus \fontdimen4\font\relax}
\providecommand{\BIBforeignlanguage}[2]{{%
\expandafter\ifx\csname l@#1\endcsname\relax
\typeout{** WARNING: IEEEtran.bst: No hyphenation pattern has been}%
\typeout{** loaded for the language `#1'. Using the pattern for}%
\typeout{** the default language instead.}%
\else
\language=\csname l@#1\endcsname
\fi
#2}}
\providecommand{\BIBdecl}{\relax}
\BIBdecl

\bibitem{imp}
P.~Zhang, H.~Luo, W.~Zhu, J.~Yang, N.~Zeng, Y.~Fan, S.~Wen, N.~Xiang, F.~Jia, and C.~Fang, ``Real-time navigation for laparoscopic hepatectomy using image fusion of preoperative 3d surgical plan and intraoperative indocyanine green fluorescence imaging,'' \emph{Surgical endoscopy}, vol.~34, pp. 3449--3459, 2020.

\bibitem{tmi}
Z.~Yang, J.~Pan, J.~Dai, Z.~Sun, and Y.~Xiao, ``Self-supervised lightweight depth estimation in endoscopy combining cnn and transformer,'' \emph{IEEE Transactions on Medical Imaging}, vol.~43, no.~5, pp. 1934--1944, 2024.

\bibitem{fm}
J.~Song, Q.~Zhu, J.~Lin, and M.~Ghaffari, ``Bdis: Bayesian dense inverse searching method for real-time stereo surgical image matching,'' \emph{IEEE Transactions on Robotics}, vol.~39, no.~2, pp. 1388--1406, 2022.

\bibitem{fm_no}
D.~Psychogyios, E.~Mazomenos, F.~Vasconcelos, and D.~Stoyanov, ``Msdesis: Multitask stereo disparity estimation and surgical instrument segmentation,'' \emph{IEEE transactions on medical imaging}, vol.~41, no.~11, pp. 3218--3230, 2022.

\bibitem{sm}
R.~Wu, P.~Liang, Y.~Liu, Y.~Huang, W.~Li, and Q.~Chang, ``Laparoscopic stereo matching using 3-dimensional fourier transform with full multi-scale features,'' \emph{Engineering Applications of Artificial Intelligence}, vol. 139, p. 109654, 2025.

\bibitem{cgi}
G.~Xu, H.~Zhou, and X.~Yang, ``Cgi-stereo: Accurate and real-time stereo matching via context and geometry interaction,'' \emph{arXiv preprint arXiv:2301.02789}, 2023.

\bibitem{sde}
V.~K. Repala and S.~R. Dubey, ``Dual cnn models for unsupervised monocular depth estimation,'' in \emph{Pattern Recognition And Machine Intelligence: 8th International Conference, PReMI 2019, Tezpur, India, December 17-20, 2019, Proceedings, Part I}.\hskip 1em plus 0.5em minus 0.4em\relax Springer, 2019, pp. 209--217.

\bibitem{sde2}
D.~Xu, W.~Wang, H.~Tang, H.~Liu, N.~Sebe, and E.~Ricci, ``Structured attention guided convolutional neural fields for monocular depth estimation,'' in \emph{Proceedings of the IEEE conference on computer vision and pattern recognition}, 2018, pp. 3917--3925.

\bibitem{sde3}
F.~Mahmood, R.~Chen, and N.~J. Durr, ``Unsupervised reverse domain adaptation for synthetic medical images via adversarial training,'' \emph{IEEE transactions on medical imaging}, vol.~37, no.~12, pp. 2572--2581, 2018.

\bibitem{sde4}
R.~J. Chen, T.~L. Bobrow, T.~Athey, F.~Mahmood, and N.~J. Durr, ``Slam endoscopy enhanced by adversarial depth prediction,'' \emph{arXiv preprint arXiv:1907.00283}, 2019.

\bibitem{monovit}
C.~Zhao, Y.~Zhang, M.~Poggi, F.~Tosi, X.~Guo, Z.~Zhu, G.~Huang, Y.~Tang, and S.~Mattoccia, ``Monovit: Self-supervised monocular depth estimation with a vision transformer,'' in \emph{2022 international conference on 3D vision (3DV)}.\hskip 1em plus 0.5em minus 0.4em\relax IEEE, 2022, pp. 668--678.

\bibitem{litemono}
N.~Zhang, F.~Nex, G.~Vosselman, and N.~Kerle, ``Lite-mono: A lightweight cnn and transformer architecture for self-supervised monocular depth estimation,'' in \emph{Proceedings of the IEEE/CVF Conference on Computer Vision and Pattern Recognition}, 2023, pp. 18\,537--18\,546.

\bibitem{monodiff}
S.~Shao, Z.~Pei, W.~Chen, D.~Sun, P.~C. Chen, and Z.~Li, ``Monodiffusion: self-supervised monocular depth estimation using diffusion model,'' \emph{IEEE Transactions on Circuits and Systems for Video Technology}, 2024.

\bibitem{afsfm}
S.~Shao, Z.~Pei, W.~Chen, W.~Zhu, X.~Wu, D.~Sun, and B.~Zhang, ``Self-supervised monocular depth and ego-motion estimation in endoscopy: Appearance flow to the rescue,'' \emph{Medical image analysis}, vol.~77, p. 102338, 2022.

\bibitem{iidsfm}
B.~Li, B.~Liu, M.~Zhu, X.~Luo, and F.~Zhou, ``Image intrinsic-based unsupervised monocular depth estimation in endoscopy,'' \emph{IEEE Journal of Biomedical and Health Informatics}, pp. 1--11, 2024.

\bibitem{endodac}
B.~Cui, M.~Islam, L.~Bai, A.~Wang, and H.~Ren, ``Endodac: Efficient adapting foundation model for self-supervised depth estimation from any endoscopic camera,'' in \emph{International Conference on Medical Image Computing and Computer-Assisted Intervention}.\hskip 1em plus 0.5em minus 0.4em\relax Springer, 2024, pp. 208--218.

\bibitem{8593623}
M.~Turan, E.~P. Ornek, N.~Ibrahimli, C.~Giracoglu, Y.~Almalioglu, M.~F. Yanik, and M.~Sitti, ``Unsupervised odometry and depth learning for endoscopic capsule robots,'' in \emph{2018 IEEE/RSJ International Conference on Intelligent Robots and Systems (IROS)}, 2018, pp. 1801--1807.

\bibitem{liu2019dense}
X.~Liu, A.~Sinha, M.~Ishii, G.~D. Hager, A.~Reiter, R.~H. Taylor, and M.~Unberath, ``Dense depth estimation in monocular endoscopy with self-supervised learning methods,'' \emph{IEEE transactions on medical imaging}, vol.~39, no.~5, pp. 1438--1447, 2019.

\bibitem{li2020unsupervised}
L.~Li, X.~Li, S.~Yang, S.~Ding, A.~Jolfaei, and X.~Zheng, ``Unsupervised-learning-based continuous depth and motion estimation with monocular endoscopy for virtual reality minimally invasive surgery,'' \emph{IEEE transactions on industrial informatics}, vol.~17, no.~6, pp. 3920--3928, 2020.

\bibitem{endoslam}
K.~B. Ozyoruk, G.~I. Gokceler, T.~L. Bobrow, G.~Coskun, K.~Incetan, Y.~Almalioglu, F.~Mahmood, E.~Curto, L.~Perdigoto, M.~Oliveira \emph{et~al.}, ``Endoslam dataset and an unsupervised monocular visual odometry and depth estimation approach for endoscopic videos,'' \emph{Medical image analysis}, vol.~71, p. 102058, 2021.

\bibitem{liu2023self}
Y.~Liu and S.~Zuo, ``Self-supervised monocular depth estimation for gastrointestinal endoscopy,'' \emph{Computer Methods and Programs in Biomedicine}, vol. 238, p. 107619, 2023.

\bibitem{da}
L.~Yang, B.~Kang, Z.~Huang, X.~Xu, J.~Feng, and H.~Zhao, ``Depth anything: Unleashing the power of large-scale unlabeled data,'' in \emph{Proceedings of the IEEE/CVF Conference on Computer Vision and Pattern Recognition}, 2024, pp. 10\,371--10\,381.

\bibitem{da2}
L.~Yang, B.~Kang, Z.~Huang, Z.~Zhao, X.~Xu, J.~Feng, and H.~Zhao, ``Depth anything v2,'' \emph{arXiv:2406.09414}, 2024.

\bibitem{dvs}
Y.~Zhou, S.~He, H.~Wang, F.~Huang, M.~Liu, Q.~Li, and Z.~Wang, ``Improved self-supervised monocular endoscopic depth estimation based on pose alignment-friendly dynamic view selection,'' in \emph{2024 IEEE International Conference on Bioinformatics and Biomedicine (BIBM)}.\hskip 1em plus 0.5em minus 0.4em\relax IEEE, 2024, pp. 3005--3012.

\bibitem{st}
M.~Jaderberg, K.~Simonyan, A.~Zisserman \emph{et~al.}, ``Spatial transformer networks,'' \emph{Advances in neural information processing systems}, vol.~28, 2015.

\bibitem{vb}
Y.~Wang, Y.~Yang, Z.~Yang, L.~Zhao, P.~Wang, and W.~Xu, ``Occlusion aware unsupervised learning of optical flow,'' in \emph{Proceedings of the IEEE conference on computer vision and pattern recognition}, 2018, pp. 4884--4893.

\bibitem{pcc}
Z.~Wang, Y.~Zhou, S.~He, T.~Li, F.~Huang, Q.~Ding, X.~Feng, M.~Liu, and Q.~Li, ``Monopcc: Photometric-invariant cycle constraint for monocular depth estimation of endoscopic images,'' \emph{arXiv preprint arXiv:2404.16571}, 2024.

\bibitem{edm}
D.~Recasens, J.~Lamarca, J.~M. F{\'a}cil, J.~M. Montiel, and J.~Civera, ``Endo-depth-and-motion: Reconstruction and tracking in endoscopic videos using depth networks and photometric constraints,'' \emph{IEEE Robotics and Automation Letters}, vol.~6, no.~4, pp. 7225--7232, 2021.

\bibitem{scared}
M.~Allan, J.~Mcleod, C.~Wang, J.~C. Rosenthal, Z.~Hu, N.~Gard, P.~Eisert, K.~X. Fu, T.~Zeffiro, W.~Xia \emph{et~al.}, ``Stereo correspondence and reconstruction of endoscopic data challenge,'' \emph{arXiv preprint arXiv:2101.01133}, 2021.

\bibitem{hamlyn}
\BIBentryALTinterwordspacing
S.~Giannarou, D.~Stoyanov, D.~Noonan, G.~Mylonas, J.~Clark, M.~Visentini-Scarzanella, P.~Mountney, and G.-Z. Yang, ``Hamlyn centre laparoscopic / endoscopic video datasets.'' [Online]. Available: \url{https://hamlyn.doc.ic.ac.uk/vision/}
\BIBentrySTDinterwordspacing

\end{thebibliography}

\end{document}